\documentclass[conference,letterpaper]{IEEEtran}
\IEEEoverridecommandlockouts
\IEEEsettopmargin{t}{72pt}
\usepackage{amsmath,amssymb,amsfonts}
\usepackage{algorithmic}
\usepackage{graphicx}
\usepackage{textcomp}
\usepackage{xcolor}

\usepackage{threeparttable}
\usepackage{ifthen}
\usepackage[hidelinks]{hyperref}
\usepackage{cleveref}

\crefname{figure}{Fig.}{Fig.}
\crefname{equation}{}{}

\newcommand{\difffrac}[2]{\frac{d#1}{d#2}}

\newcommand{\transpose}[1]{#1^T}
\newcommand{\covd}[1]{\nabla_{#1}}
\newcommand{\dotprod}[2]{\langle  #1 , #2 \rangle}
\newcommand{\covprod}[2]{\langle  #1 ; #2 \rangle}
\newcommand{\distfunc}[2]{#1(#2,#2)}
\newcommand{\distfunct}[3]{#1(#2,#3)}

\newcommand{\norm}[1]{\left\lVert #1 \right\rVert}
\newcommand{\inv}[1]{#1^{\text{-}1}}

\newcommand{\shape}{r}
\newcommand{\dshape}{\dot{\shape}}
\newcommand{\ddshape}{\ddot{\shape}}
\newcommand{\gait}{\phi}

\newcommand{\power}{P}
\newcommand{\mass}{M}
\newcommand{\corio}{C}
\newcommand{\drag}{D}
\newcommand{\force}{F}
\newcommand{\covyank}[1][]{
    \ifthenelse{\equal{#1}{}}
    {
        {\covd{\dgcurve}\effort}_{\gcurve}%
    }
    {
        \left({\covd{\dgcurve}\effort}\right)_{#1}%
    }
}
\newcommand{\covtug}[1][]{
    \ifthenelse{\equal{#1}{}}
    {
        {\covd{\dgcurve}^2\effort}_{\gcurve}%
    }
    {
        \left({\covd{\dgcurve}^2\effort}\right)_{#1}%
    }
}
\newcommand{\covacc}[1][]{
    \ifthenelse{\equal{#1}{}}
        {
            a_{\gcurve}%
        }
        {
            a_{\gcurve,#1}%
        }
}
\newcommand{\covjerk}[1][]{
\ifthenelse{\equal{#1}{}}
    {
        {\covd{\dgcurve}\covacc}_{\gcurve}%
    }
    {
        \left({\covd{\dgcurve}\covacc}\right)_{#1}%
    }
}
\newcommand{\covsnap}[1][]{
    \ifthenelse{\equal{#1}{}}
    {
        {\covd{\dgcurve}^2\covacc}_{\gcurve}%
    }
    {
        \left({\covd{\dgcurve}^2\covacc}\right)_{#1}%
    }
}
\newcommand{\effort}{E}

\newcommand{\currentgait}{\gait_0}
\newcommand{\dcurrentgait}{\dot{\gait}_0}
\newcommand{\targetgait}{\gait_d}
\newcommand{\dtargetgait}{\dot{\gait}_d}
\newcommand{\currentgaitpar}{t_0}
\newcommand{\targetgaitpar}{t_f}

\newcommand{\gcurve}[1][]{\gamma}
\newcommand{\dgcurve}[1][]{\dot{\gcurve}}
\newcommand{\ddgcurve}[1][]{\ddot{\gcurve}}
\newcommand{\metric}{g}
\newcommand{\dual}[1]{{#1}^{*}}
\newcommand{\dualmetric}{\dual{\metric}}
\newcommand{\cometric}{\widetilde{\metric}}
\newcommand{\inmetric}{h}
\newcommand{\varfam}{\Gamma}
\newcommand{\varvar}{s}
\newcommand{\varfield}{V}

\newcommand{\RCurva}{R}

\begin{document}
\title{Optimal Control Approach for Gait Transition with Riemannian Splines
\thanks{This work was supported in part by the Office of Naval Research under awards N00014-23-1-2171.}
}

\author{Jinwoo Choi and Ross L. Hatton 
    \thanks{Jinwoo Choi and Ross L. Hatton are with the Collaborative Robotics and Intelligent Systems (CoRIS) Institute at Oregon State University, Corvallis, OR USA. \{\href{mailto:choijinw@oregonstate.edu}{choijinw}, \href{mailto:Ross.Hatton@oregonstate.edu}{Ross.Hatton}\}@oregonstate.edu}
}

\maketitle

\begin{abstract}
Robotic locomotion often relies on sequenced gaits to efficiently convert control input into desired motion. Despite extensive studies on gait optimization, achieving smooth and efficient gait transitions remains challenging. In this paper, we propose a general solver based on geometric optimal control methods, leveraging insights from previous works on gait efficiency. Building upon our previous work, we express the effort to execute the trajectory as distinct geometric objects, transforming the optimization problems into boundary value problems. To validate our approach, we generate gait transition trajectories for three-link swimmers across various fluid environments. This work provides insights into optimal trajectory geometries and mechanical considerations for robotic locomotion.
\end{abstract}

\begin{IEEEkeywords}
robotics, nonholonomic systems, optimal control
\end{IEEEkeywords}

\section{Introduction}

\begin{figure}[ht]
    \centering
    \includegraphics[width = 0.4\textwidth]{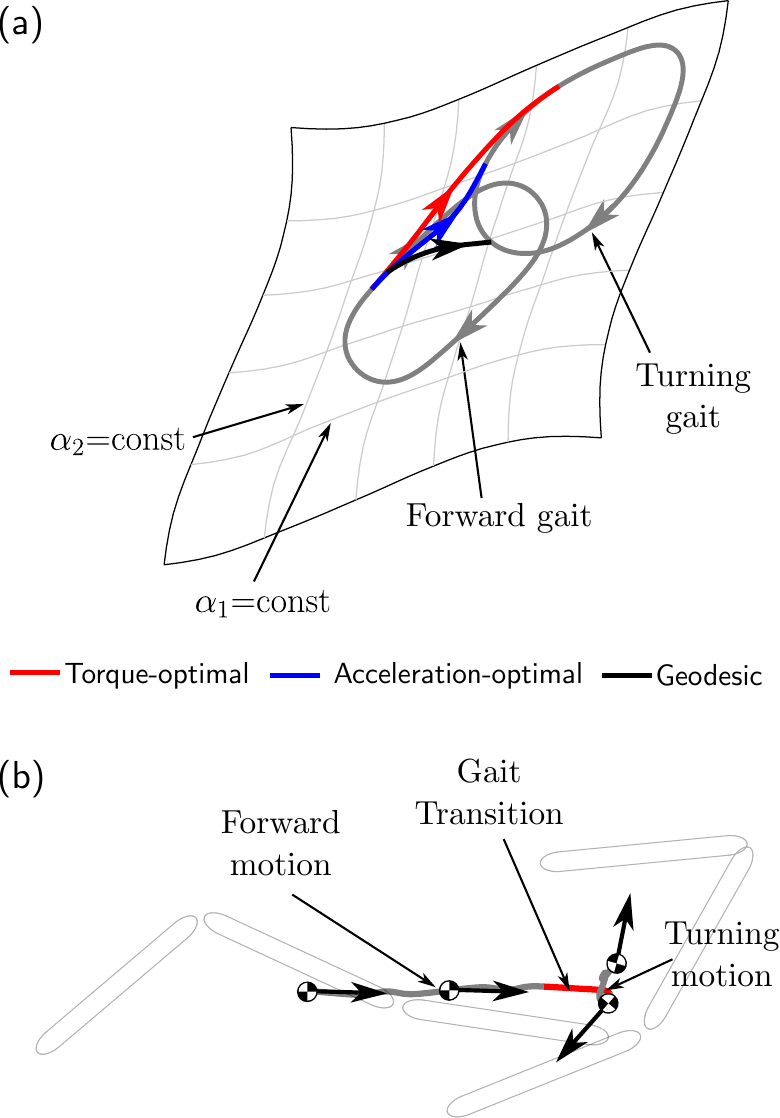}
    \caption{The optimal gait transition for inertia-dominated systems from the gait generating the forward motion to the gait generating the turning motion. (a) The three different gait-switching trajectories starting from the same point in metric-weighted space. $\alpha_1$ and $\alpha_2$ are the joint angles of the swimmer. The red curve denotes the torque-optimal trajectory, the blue curve denotes the acceleration-optimal trajectory, and the black curve denotes the geodesic trajectory. The geodesic trajectory, the shortest curve between points, nonsmoothly connects to each gait. The torque-optimal and acceleration-optimal trajectories smoothly connect to each gait. (b) The body trajectory generated by the gait switching scenario: a sequence consisting of the forward gait, the torque-optimal gait transition, and the turning gait. The center of mass markers and black arrows denote where the swimmer's head is pointing at each time step. The first gray, red, and the second gray curve respectively indicate the body trajectory generated by the forward gait, the torque-optimal gait transition, and the turning gait.}
    \label{fig:graphical_abstract}
\end{figure}

Robotic locomoting systems move by using constraints and other interactions with their environments to convert internal shape changes into motion through the world. For robots whose joint motion is bounded (as compared to the continuous rotation afforded by wheels and propellers), it is often useful to focus attention on gaits---cyclic shape inputs that produce characteristic net displacements and individually remain within the joint limits. In previous work, we and others have explored how to find gaits that most efficiently convert a system’s input effort into forward, lateral, and turning motion \cite{tam2007optimal,ramasamy2019geometry,alouges2019energyoptimal,hatton2022geometry}, and how to generate continuous gait families parameterized by quantities such as steering rate or step size, in which each gait is optimally efficient out of all gaits that produce motion in the same direction and with the same step size \cite{deng2022enhancing, choi2022optimal}.

By sequencing these motion primitives, a gait-based controller can focus on how to navigate to a goal point while guaranteeing that joint motions remain within their limits and are locally optimal with respect to energy costs. One difficulty with sequenced-gait control is that even though repeated execution of one gait cycle makes robots move smoothly and efficiently, sequencing of gaits among the gait families might not guarantee optimality and smoothness. Furthermore, the gait-transition trajectory could generate unexpected motion results acting as a disturbance of the motion control. Thus, a full gait-based controller should include gait-transition trajectories with the following characteristics:
\begin{enumerate}
    \item Smoothness and continuity during gait transition sequences.
    \item High efficiency in terms of the associated effort.
    \item Small disturbance or execution time over the gait transition.
\end{enumerate}

In this paper, we identify the geometric structure of the optimal gait transition problem and propose a general solver to this problem, based on a geometric optimal control approach presented in \cite{cabrera2024optimal}. From our perspective, the associated cost can be interpreted as certain geometric properties of the trajectory. For example, the dissipated power for drag-dominated systems \cite{ramasamy2019geometry} is the metric-weighted pathlength of the trajectory, which is minimized by the geodesic as the shortest curve in the manifold (e.g., the black curve in \cref{fig:graphical_abstract}(a)). Beyond the shortest path, a smoothly connected trajectory is derived by minimizing the covariant acceleration norm (e.g., the blue curve in \cref{fig:graphical_abstract}(a)). Finally, the covariant acceleration norm can represent the necessary actuator torque to achieve the corresponding motion by selecting the cometric that matches the physical configuration of the actuator \cite{hatton2022geometry} (e.g., the red curve in \cref{fig:graphical_abstract}(a)). 

The trajectory optimization problem can be replaced by a system of differential equations, based on a variational approach. The structure of the differential equation is identical or similar to the geodesic equation or the spline equation \cite{noakes1989cubic,crouch1991geometry,cabrera2024optimal}. To generate the trajectory satisfying the specific condition such as an initial and final condition, it finally reduces to the boundary value problem. To solve this problem, we use an indirect shooting method. The optimization solver shoots the trajectory iteratively and finds one that minimizes the boundary value error.

To demonstrate our approach, we generate gait transition trajectories from a forward motion gait to a turning motion gait for two different three-link swimmers: swimmers immersed in a viscous and perfect fluid environment. For the viscous three-link swimmer, we generate the gait-transition trajectory based on the power cost dissipated by the surrounding medium. For the perfect-fluid three-link swimmer, we generate optimal transition trajectories using both the covariant acceleration norm of the trajectory and the actuator torque norm as the costs. As illustrated in \cref{fig:graphical_abstract}, the gait transition trajectories generated by the optimal control approach are thus solutions to connect one gait with the other gait smoothly and minimize the associated cost.

\section{Prior works}
The robotics locomotion community has exploited motion primitives in control and motion planning and paid attention to the gait transition algorithm. One approach to achieve the smooth gait transition is using the interpolation between the current gait and the desired gait \cite{gehring2013control,powell2013speed,da20162d}. This approach can guarantee the smoothness of motion but the resulting trajectory is likely to be suboptimal. 
An optimal control approach has been paid attention to most robotic locomotion communities because of generality. It is useful for not only gait optimization but also gait transition problems \cite{wensing2024optimizationbased}. Despite its versatility, it requires a high computational cost.

When a command gait is given, the current position is considered as a disturbance. Then, the gait transition problem can be thought of as a stabilization problem. A limit cycle attractor has been utilized to stabilize disturbances in previous works \cite{ijspeert2008central,seo2010cPGbased,manchester2011regions}. These works have focused on synthesizing a limit cycle oscillator to stabilize the system and execute the gait cycle. By controlling the parameters of a limit cycle (a gait pattern) oscillator, the controller induces the locomoting system to achieve the desired motion and make the system return to the desired gait pattern at the gait transition phase. It can generate a smooth modulation of several gait patterns by a few parameters and reduce the complexity of the gait transition problem. However, the optimal design of a limit cycle attractor (e.g., minimizing the gait transition cost, or designing a limit cycle to follow efficient gaits) is challenging. 

\section{Geometric Mechanics Framework}
\subsection{Curved space}
In this subsection, we briefly review \emph{Riemannian Geometry} from the perspective of optimal control problems to understand how the optimal trajectory can be interpreted geometrically. The detailed formulations or definitions of notions can be found in \cite{lee1997riemannian}.

A Riemannian \emph{metric} $\metric$ for the space contains the distance information and is defined by the inner product of each basis. With the metric, a dot product on the Riemannian manifold is defined as 
\begin{equation}
    \metric\left(X,Y\right) = \transpose{X}MY,
\end{equation}
where $X$ and $Y$ are arbitrary tangent vectors on the Riemannian manifold, $\metric$ is the Riemannian metric, and $\mass$ is the metric tensor associated with the Riemannian manifold.

A \emph{geodesic} is a structure corresponding to the notion of a ``straight line'' in Euclidean space, in that both are the shortest paths between one point and the other. If we think of a particle with a constant speed or zero acceleration in a flat space, the particle generates a straight line. Similarly, the corresponding trajectory in a curved space can be derived by finding a curve with a zero \emph{covariant acceleration}. Thus, a curve $\gcurve$ is a geodesic if it satisfies
\begin{equation}
    \covacc = \covd{\dgcurve}\dgcurve = 0,
    \label{eq:geoddef}
\end{equation}
where $\gcurve(t)$ is a time-parameterized curve, $\covacc$ is a covariant acceleration, $\covd{}$ is the Levi-Civita Connection that is compatible with a metric $\metric$, and $\covd{X}Y$ is a \emph{covariant derivative} of a tensor (vector) field $Y$ along a tensor (vector) field $X$ which provides tool to differentiate a tensor field as if it was on the true geometry of the manifold, correcting for any coordinate-induced distortions. Unlike a flat space, the local coordinate bases of a curved space could vary at each point. A covariant derivative includes the correction terms of changes in the local coordinate bases. Thus, there is no acceleration along a geodesic curve except for an acceleration resulting from a curved space.

\subsection{Locomoting systems}
\label{subsec:locomoting}
\begin{figure*}
    \centering
    \includegraphics[width=0.9\textwidth]{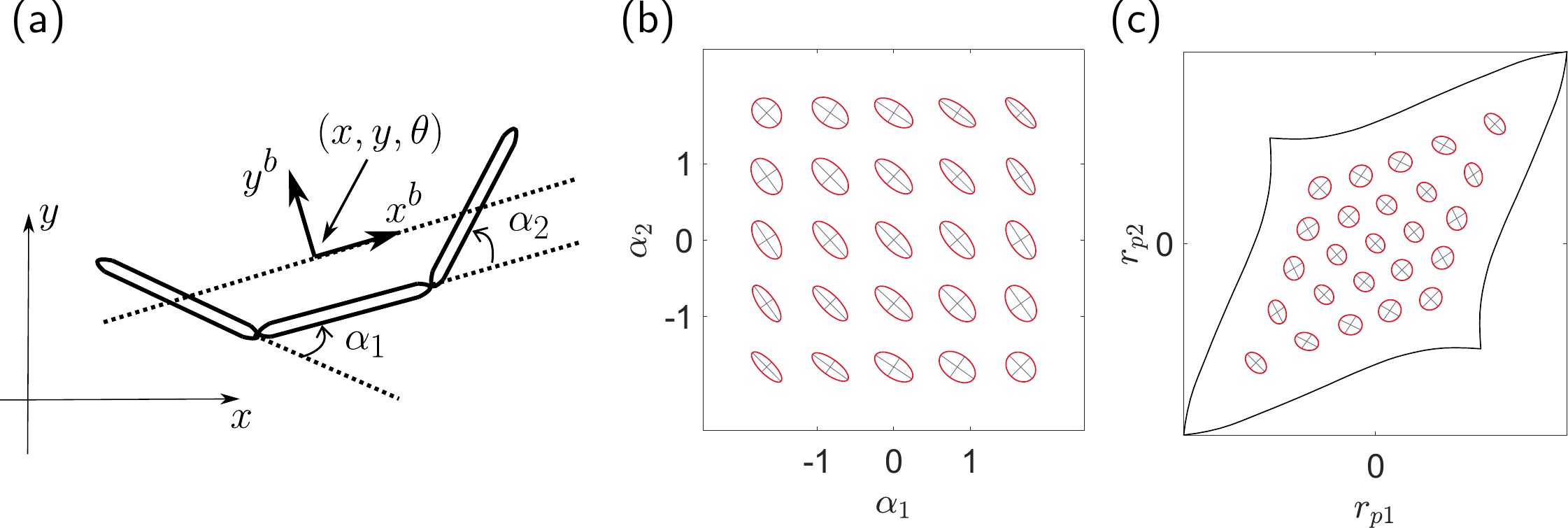}
    \caption{(a) Schematic of a three-link viscous swimmer. The shape of the system is described by two joint angles and is a two-dimensional shape space. $\alpha$ denotes the joint angles. $x$-$y^b$ denotes the body coordinate of the system. (b) Visualization of the metric field of viscous three-link swimmers. Infinitesimal circles at each point indicate how much the shape space distorts in each direction in comparison to the actual space. Physically, it means how much effort is required to produce shape changes in a given direction at each point. Longer ellipse axes denote directions in which shape changes are cheaper. (c) The projection of the actual space onto the flat 2D plane that minimizes distortion. Because the actual space is curved, the approximation must distort when forced into the flat plane. there are still some elliptically of the infinitesimal circles. The pathlength in this manifold can approximate the actual effort to change from one shape to another shape.}
    \label{fig:systembasismetric}
\end{figure*}
In this paper, we use three-link swimmers in viscosity- and inertia-dominated fluids in \cref{fig:systembasismetric} as example systems to demonstrate the efficacy of the proposed gait-switching algorithm. Due to the two-dimensional shape space, it is easy to visualize their shape trajectory over the gait transition and how their body motion changes. Even though we use two-dimensional locomotion systems to demonstrate the optimal control approach, this approach applies to high-dimensional systems \cite{cabrera2024optimal}.

Locomoting systems generate their body motion by the interaction between the environment and their internal shape change. As our example system, three-link swimmer's shape spaces are usually parameterized by their joint angles, $r=(\alpha_1,\alpha_2) \in \mathbb{S}^2$. The position is defined as the specially chosen body frame with respect to the world frame.\footnote{We choose the body frame to minimize the parallel-parking motion effect generated when we integrate the body velocity. \cite{hatton2015Nonconservativity}}

From the perspective of the optimal control problem, a measure of the cost is generally represented by the integral of the instantaneous cost (or running cost). The instantaneous cost could be defined by several criteria such as penalizing undesired state, minimizing tracking error, or minimizing effort. For the purpose of this paper, we narrow down the sense of the cost as the effort needed to execute the gait or the trajectory.

\subsubsection{Drag-dominated systems}
For viscous (drag-dominated) systems, the equations of motion are first order, with the force applied to the system equal to the product of the generalized drag tensor and the system’s velocity,
\begin{equation}
    \force(t) = \drag(\shape) \dshape,
    \label{eq:viscousmotion}
\end{equation}
where $\force$ is the actuator force exerting on joints, and $\drag(\shape)$ is the generalized drag metric tensor.\footnote{The formal derivation of the generalized drag metric tensor $\drag(\shape)$ for viscous three-link swimmers can be found in \cite{ramasamy2019geometry}.} 

The same amount of joint action in one direction could need more effort than that in another direction. For example, the symmetric joint motion of a viscous swimmer, along $\shape_1 = \shape_2$, is more expensive than the antisymmetric joint motion, along $\shape_1 = -\shape_2$.\footnote{The fluid drag heavily depends on the area of the object facing the fluid. Symmetric joint motion induces substantial lateral motion in all three links of the swimmer. In contrast, the antisymmetric joint motion induces substantial longitudinal motion.} Suppose that the shape space is the parameterization of the actual space which considers the effort to change the shape of locomoting systems physically. Then, the pathlength of a shape trajectory in the actual effort space tells how much effort is needed when the system executes a shape trajectory. A geodesic can be interpreted as an optimal shape trajectory between one point and the other.

From this perspective, the generalized drag metric tensor defines the Riemannian metric $\metric$ to contain the distance information of each direction in the shape space, and the pathlength of the shape trajectory, $\gcurve(t)$, can be interpreted as the power, $\power(\gcurve)$, dissipated into the surrounding medium when the system executes the trajectory \cite{hatton2017kinematic},
\begin{equation}
    \power(\gcurve) = \int_{t_0}^{t_f} \sqrt{\distfunc{\metric}{\dgcurve}} dt = \int_{t_0}^{t_f} \norm{\dgcurve}_{\metric} dt,
    \label{eq:distcost}
\end{equation}
where $\distfunc{\metric}{\cdot}$ is the inner product of two vectors under the metric $\metric$. As another cost of drag-dominate systems, the actuator torque dissipated into the surrounding medium can be considered. In this case, the squared generalized drag metric tensor defines the Riemannian metric, and the path length in this Riemannian manifold measures the squared actuator torque.

\subsubsection{Inertia-dominated systems}
For the perfect-fluid (inertia-dominated) system, the metric tensor describes the inertia (or generalized mass) information for each axis of the shape space  \cite{hatton2022geometry}, and provides an equation of motion for an inertia-dominated system,
\begin{equation}
    \force(t) = \mass(\shape) \ddshape + \corio(\shape,\dshape),
    \label{eq:inertiamotion}
\end{equation}
where $\mass(\shape)$ is the generalized mass metric tensor,\footnote{The formal derivation of the generalized mass metric tensor $\mass(\shape)$ for perfect-fluid three-link swimmers can be found in \cite{hatton2022geometry,kanso2005locomotion}.} $\corio(\shape,\dshape)$ is generated from a derivative of $\mass$ and encodes the centrifugal and Coriolis forces acting on the system, and $\force$ is the actuator force exerting on joints. The acceleration dictated by the external force is described by the covariant acceleration along a trajectory $\gcurve$,
\begin{equation}
    \covacc = \covd{\dgcurve}\dgcurve = \ddshape + \inv{\mass}\corio.
\end{equation}
Note that the terms, $\inv{\mass}\corio$, can be interpreted as the correction term in the covariant derivative and are also known as Christoffel symbols \cite{hatton2022geometry}.

A \emph{dual metric}, $\dualmetric$, can be used to define the acceleration-based cost. The relationship between metric tensors and dual-metric tensors is the inverse of each other, $\mass^{*} = \mass^{-1}$. With this choice of cometric, the norms of force covectors\footnote{A covector is a linear function that maps a vector to a scalar value. It is natural to define force as a covector because power is a scalar value that is the dot product of force (covector) and velocity (vector).} induce the norms of the covariant acceleration, which tells how much the total force is needed to move a particle to follow the trajectory,
\begin{equation}
     \norm{\force}_{\dualmetric} = \norm{\covacc}_{\metric}.
\end{equation}
The squared norm of the actuator torque could be used as one of the costs commonly used in classical optimal control problems. The force covector in \eqref{eq:inertiamotion} to generate the acceleration $\covacc$ is described by the product of the mass metric tensor $\mass$ and the covariant acceleration $\covacc$. Thus, we introduce a \emph{torque-cost cometric}, $\cometric$, which considers the physical configuration of actuators. If the actuator coordinate is aligned with the shape coordinate physically, the cometric tensor reduces to a diagonal matrix. With this choice of cometric, the norms of forces can capture the actual actuator effort. We define an induced metric so that the acceleration norm associated with the induced metric is equivalent to the torque-effort,
\begin{equation}
     \norm{\force}_{\cometric} = \norm{\covacc}_{\inmetric}.
\end{equation}

\subsection{Variational problem}
\label{subsec:variation}

\begin{figure}
    \centering
    \includegraphics[width=0.5\textwidth]{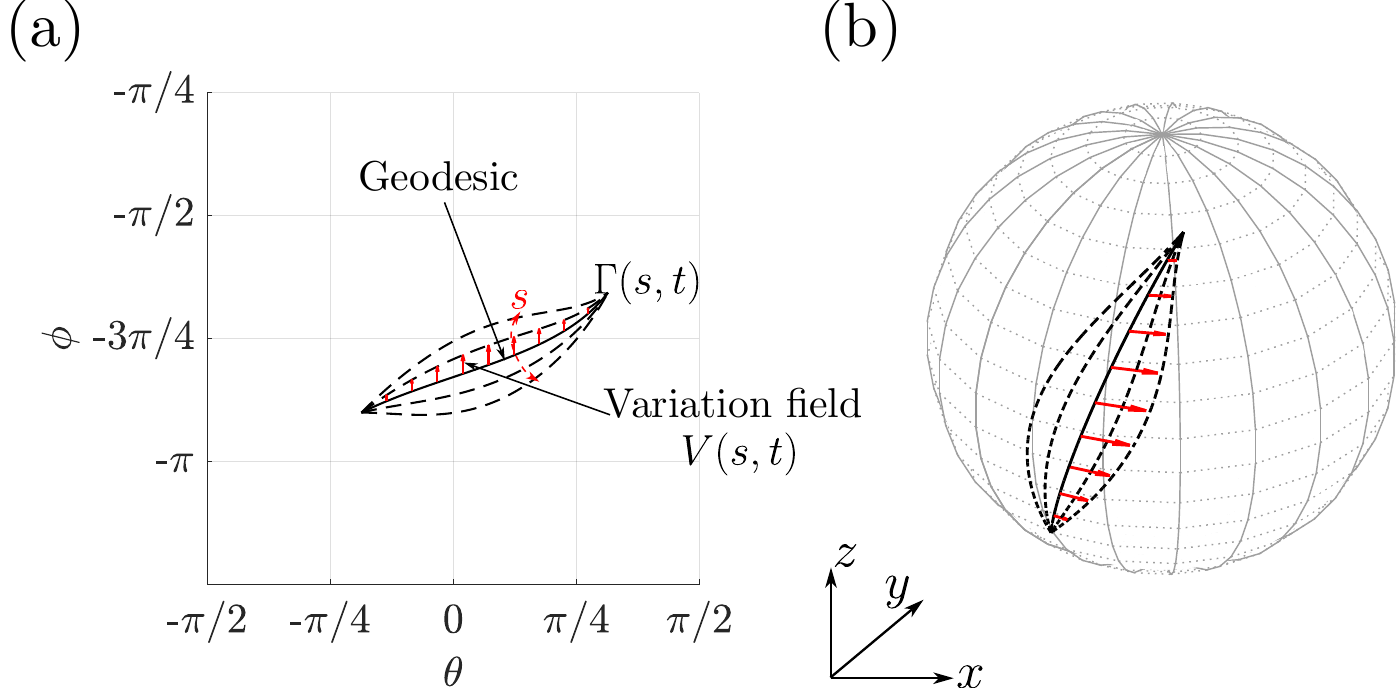}
    \caption{The variational curves (a) in the spherical parameter space (b) and the spherical space, respectively. The spherical manifold is parameterized by the polar angle $\theta \in [-\pi/2,\pi/2]$ and the azimuthal angle $\phi \in [-\pi,\pi]$. Suppose that the variation family $\varfam(\varvar,t)$ is parameterized by one parameter $\varvar$, and the curves in the family have the same initial and final points. The black curves are in the same variational family. The black solid curve is a geodesic. The shortest path among curves in the family is a geodesic curve, and the covariant acceleration is zero along the geodesic curve. The red arrow indicates the variation field, $\varfield(\varvar,t)$, which tells the infinitesimal difference between the neighborhood of the geodesic.}
    \label{fig:spherical_space}
\end{figure}

The geometric optimal control problem can be seen as a variational problem to find the maxima or minima of the cost function. The cost functions are mostly defined as integrals of the geometric property of the curve. Here, we briefly review the spline equation corresponding to the variational problem we are interested in. The detailed derivation of the spline equations is presented in \cite{balseiro2017simple,cabrera2024optimal}.

Suppose that $\varfam(\varvar,t)$ is a smooth one-parameter variation of $\gcurve(t)$. As shown in \cref{fig:spherical_space}, a variation field $\varfield(\varvar,t)$ is a vector field along the curve $\gcurve$ that indicates the difference between neighbor curves in a variation family. Suppose that variation of curves $\varfam$ is \emph{proper} variation\footnote{The curves in proper variations have the same initial and final points.}, and the curve has a unit speed. Then, a derivative of a pathlength over the variation family \cite{lee1997riemannian} is
\begin{equation}
    \difffrac{}{\varvar}\power(\varfam) = -\int_{0}^{T}\distfunct{\metric}{\varfield}{\covd{\dgcurve}\dgcurve}.
    \label{eq:pathlenvar}
\end{equation}
If $\gcurve$ is the curve minimizing a pathlength, the derivative of a pathlength vanishes. It means that every curve minimizing the pathlength is a geodesic.

We define the effort covector $\effort$ as the induced-metric dual of the metric-covariant acceleration. The variation of the effort norm squared associated \cite{crouch1991geometry} is
\begin{multline}
    \frac{1}{2}\difffrac{}{\varvar}\int_{0}^{T}\norm{\covacc}_{\inmetric}^2dt= 
    \frac{1}{2}\difffrac{}{\varvar}\int_{0}^{T}\norm{\effort}_{\dual{\inmetric}}^2dt=\\ \int_{0}^{T}\dotprod{\varfield}{\covd{\dgcurve}^2\effort + \covprod{\RCurva_{\metric}(\bullet,\dgcurve)\dgcurve}{\effort}}-\frac{1}{2}\nabla_V\dual{\inmetric}\left(\effort,\effort\right),
\end{multline}
where $\RCurva_{\metric}$ is the Riemannian curvature tensor, which describes how much “free” velocity change the trajectory can capture by exploiting the curvature of the manifold, $\covprod{\cdot}{\cdot}$ is a tensor contraction operator, $\bullet$ indicates an element of the dual vector space obtained by contraction, and $\nabla_\bullet\inmetric^*$ measures the incompatibility of the dual induced metric $\dual{\inmetric}$ with respect to the inertia metric geometry $\metric$ \cite{cabrera2024optimal}. Given arbitrary vector field $\varfield$, every curve minimizing the effort norm squared should satisfy
\begin{equation}
    \covd{\dgcurve}^2\effort + \covprod{\effort}{\RCurva_\metric(\bullet,\dgcurve)\dgcurve} - \frac{1}{2}\nabla_{\bullet}\inmetric^*(\effort,\effort) = 0.
    \label{eq:effspline}
\end{equation}
For the torque-optimal trajectory, we choose the induced metric so that the effort norm represents the actuator-torque norm,
\begin{equation}
    \norm{\force}_{\cometric} = \norm{\effort}_{\dual{\inmetric}}.
\end{equation}
For the acceleration-optimal trajectory, we choose the mass metric $\metric$ as the induced metric \inmetric so that the effort norm is equal to the covariant acceleration norm, the equation reduces to   
\begin{equation}
    \covd{\dgcurve}^2\covacc + \RCurva_{\metric}(\covacc,\dgcurve)\dgcurve = 0.
    \label{eq:accspline}
\end{equation}
\vspace{-5mm}
\section{Optimal Gait Transition}
We can pose the optimization problem in terms of 
\begin{enumerate}
    \item finding the shortest path (or minimal energy functional) from the current point to a point on the new gait,
    \begin{equation}
        \min_{\gcurve,T}\int_0^T \norm{\dgcurve}_{\metric}^2dt,
    \end{equation}
    \item or by finding the trajectory that synchronizes with the new gait while incurring the least acceleration away from the system’s natural trajectories,    
    \begin{equation}
        \min_{\gcurve,T}\int_0^T \norm{\covacc}_{\metric}^2dt,
    \end{equation}
    \item or the least actuator force (where the difference between the acceleration- and force-minimizing trajectories is that the latter accounts for the extra force incurred when the motors act antagonistically),
    \begin{equation}
        \min_{\gcurve,T}\int_0^T \norm{\force}_{\cometric}^2dt,
    \end{equation}
\end{enumerate}
where $T$ is the time to execute the gait transition trajectory. To find the power-optimal trajectory of the drag-dominated systems, we use the first optimization formulation, and its solution is a geodesic. For the inertia-dominated systems, the second and third optimization problems are used, and their solutions are expressed as the spline equations in \cref{eq:effspline,eq:accspline}.

These optimization problems are transformed into a boundary value problem (BVP). To solve BVPs, we use a shooting method that continues to shoot trajectories with different initial conditions until finding the solution. We formulate the optimization problems to minimize the boundary value error term so that the solver finds the trajectory in which the final state (smoothly)\footnote{It depends on whether the boundary velocity conditions exist or not.} is connected with the gait cycles. Although splines (or geodesics), solutions to the differential equation, are locally optimal for all states, there could be multiple splines. Thus, we include the trajectory cost in the problem cost, which helps us select the global solution. The optimization problem is formulated as 
\begin{subequations}
    \label{eq:bvpopt}
\begin{align}    
    \label{subeq:bvpoptobj}
    \min_{x} \ &  \text{(BVP error)} + \text{(Trajectory Cost)}, \\    
    \label{subeq:bvpdopteq}
    \text{subject to} \ & \text{(Differential equations),} \\
    \label{subeq:bvpoptinit}
    \ &  \text{(Initial conditions)},
\end{align}
\end{subequations}
where $x$ are decision variables. As mentioned in \cref{subsec:variation}, the differential (-algebraic) equations describe the optimality condition for the trajectory. The detailed formulations of each BVP are described in the Appendix.

To demonstrate the efficacy of the proposed algorithm in this paper, we scope down scenarios to the gait transition from forward gaits to turning gaits. We generate path-optimal trajectories for viscous three-link swimmers, and acceleration-optimal and torque-optimal trajectories for perfect fluid three-link swimmers. Each swimmer has a link unit length of 10 times the link width.

\begin{figure*}[ht]
    \centering
    \includegraphics[width=\textwidth]{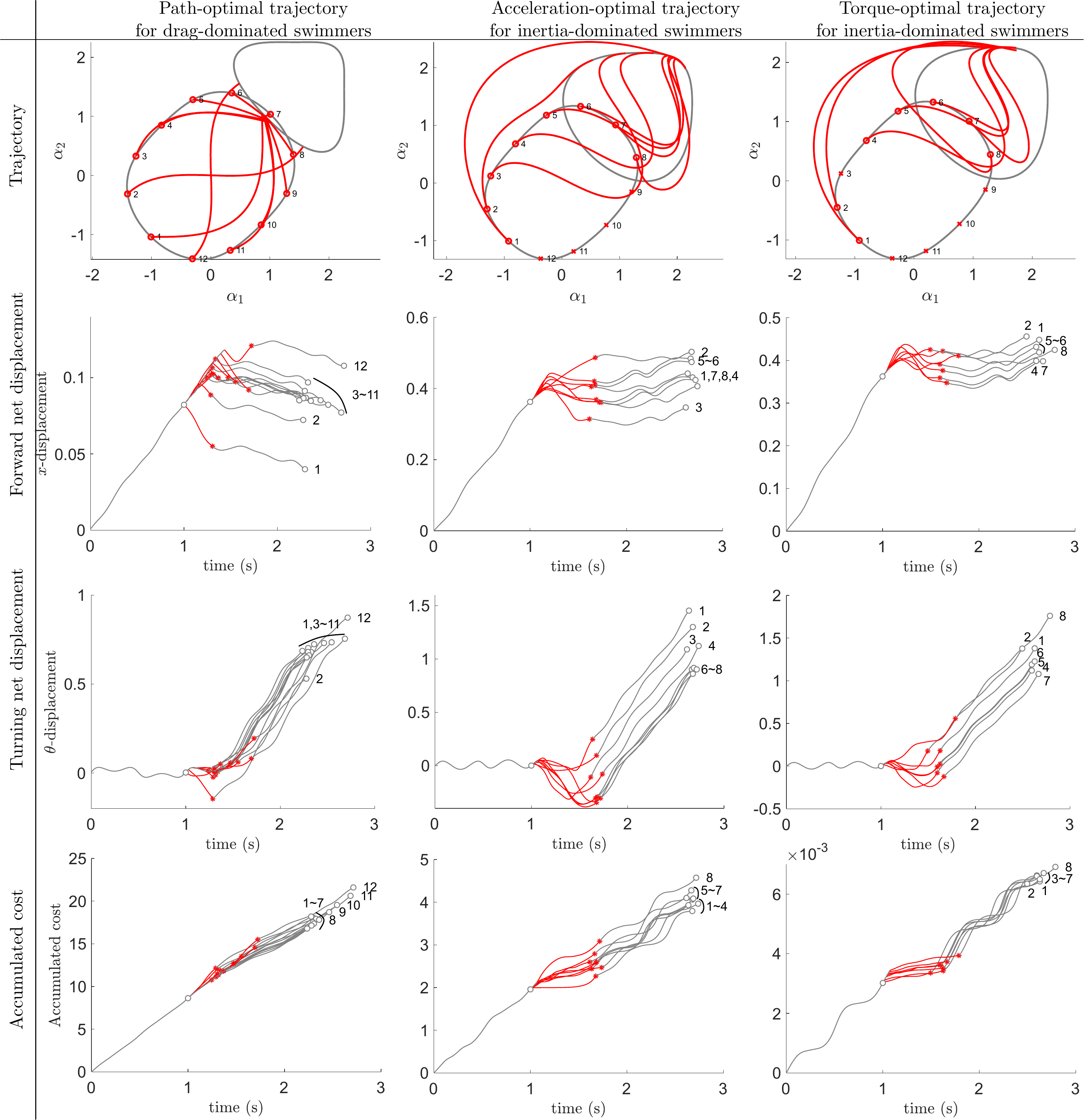}
    \caption{The gait switching trajectories from the forward gait to the turning gait. The results of columns respectively are distinguished by path-optimal trajectories for viscous swimmers, acceleration-optimal trajectories for perfect fluid swimmers, and torque-optimal trajectories for perfect fluid swimmers in order. The gait-switching scenario includes a one-cycle and extra execution of forward gait (the first gray curves), gait transition (red curves), and a one-cycle execution of turning gait (the second gray curves). Especially, the lower-left and upper-right gray closed curve in the first row of figures denotes the forward gait and turning gait, respectively. The velocity directions of both gaits are clockwise. The figures in the first row represent the shape trajectory. The second row shows the changes in forward net displacement for a single scenario, while the third row illustrates the changes in turning net displacement. Finally, the fourth row displays the changes in accumulated cost. The length unit in the figure is the link length of the swimmers. The red curves denote the gait-switching trajectories. Each marker in the first row on the forward gait denotes where the initial points are. The circle marker in the first row denotes that the solver generates the corresponding trajectory, and the diagonal cross marker in the first row indicates that the solver fails to generate the corresponding trajectory. The gait-switching trajectories start from the forward gait and join into the target gait smoothly. From the second row of figures, the numbers in the figure distinguish the trajectories based on different initial points. The red star marker tells when the execution of the gait-switching trajectory is ended. The gray circle marker tells when the execution of the forward gait or the turning gait is ended.}
    \label{fig:results}
\end{figure*}

\cref{fig:results} shows the results of the gait transition trajectories, the net displacement generated by the trajectories, and the cost to execute the trajectories. During the gait transition scenario, the process begins at point 1 and executes one full cycle of the forward gait. It continues on the forward gait until it reaches each initial point, then executes the gait transition trajectories, and finally completes one cycle of the turning gait sequentially. We picked 12 initial points to generate the trajectories.
\begin{enumerate}
    \item Path-optimal trajectories: These trajectories represent the shortest path between the initial points along the forward gait and the turning gait, and the trajectories' speeds are constant. The whole trajectories (gait-transition-gait) are piecewise nonsmooth because the optimization problems do not consider the initial and final velocity conditions. The transition cost and unexpected motions decrease as the initial point along the forward gait approaches the turning gait. Even though the trajectory starting from point 7 is relatively shorter than others, it gives a delay to the control system until reaching these points. These trajectories can be particularly useful for drag-dominated systems commanded by the velocity.
    \item Acceleration-optimal trajectories: These trajectories are designed to minimize the norm of covariant acceleration. The speed profiles for the entire trajectories, including gaits, are smooth. The acceleration-optimal trajectories ease out of and into the gait trajectories, with the cost of curvature dictated by the mass metric. Mechanically, these trajectories make the controller take advantage of effects (such as centrifugal and Coriolis forces) as much as possible to move the locomoting system.
    \item Torque-optimal trajectories: These trajectories are designed to minimize the actuator torque norm and are similar to the acceleration-optimal trajectories. However, they additionally avoid motions in which the motors would back-drive each other, at the expense of incurring more extraneous position motion. Similar to the acceleration-optimal trajectories, they ease out of and into the gait trajectories, with the cost of curvature dictated by the square of the mass metric.
\end{enumerate}

At some initial points, the solver cannot generate the trajectory because some solutions violate the joint limit or the acceleration limit that the actuator can generate. We have not implemented a solver for the optimal solutions in that case, but that we expect them to accommodate the constraints by a combination of following the boundary of the constrained region and accepting more aggressive departure and arrival accelerations.

\section{Conclusion}
In this paper, we identified a geometric structure of the optimal gait transition problem by the optimal control framework presented in \cite{cabrera2024optimal}. We demonstrated this gait transition method on drag- and inertia-dominated three-link swimmers. In comparing the several definitions of the cost, this work highlights how each optimal trajectory differs from each other to take advantage of its mechanical properties.

A line of future work is investigating how to apply this gait transition problem to a variety of locomotion systems such as high-dimensional locomotion systems \cite{yang2024geometric} or legged robots affected by a gravity force. In particular, a gravitational force can be described as a gradient of potential energy with respect to the metric \cite{cortesmonforte2002geometric}. 

We believe that this work establishes one of the important steps for the full gait-based control framework by understanding how to smoothly transform one gait into another gait. Based on our previous works, the collection of gaits parameterized by the steering rate and gait speed can be generated to reduce the complexity of controlling locomoting systems. By combining our work with the conventional trajectory tracking control of car-like robots, we expect to achieve a full controller of general locomoting systems.

\appendix[Boundary Value Problem Formulation]
\vspace{-5mm}
\begin{table}[ht]
    \centering
    \caption{Parameters of boundary value problem}
    \begin{tabular}{r|l}
        \hline
        \hline
        $\currentgait$ & function of time to describe current gait\\
        $\targetgait$ &  function of time to describe target gait \\
        $\currentgaitpar$ & time parameter of current gait \\
        $\targetgaitpar$ & time parameter of target gait \\
        $T$ & time to execute the trajectory \\
        \hline
        \hline
    \end{tabular}
    \label{tab:bvpparam}
\end{table}
\vspace{-5mm}
\subsection{Path-optimal trajectory}
\begin{enumerate}
    \item Decision variables: $\targetgaitpar, T$, and $\dgcurve_0$.
    \item Boundary value error: 
    \begin{equation}
        \norm{\gcurve(T) - \targetgait(\targetgaitpar)},
    \end{equation}
    \item Initial conditions: This constraint forces an initial point of a gait-switching trajectory should start from the current gait cycle,
    \begin{equation}
        \gcurve(0) = \currentgait(\currentgaitpar) \text{ and } \dgcurve(0) = \dgcurve_0.
    \end{equation}
\end{enumerate}
\subsection{Acceleration-optimal trajectory}
\begin{enumerate}
    \item Decision variables: $\targetgaitpar$, $T$, $\covacc[0]$, and $\covjerk[0]$.
    \item Boundary value error:
    \begin{equation}
        \norm{\gcurve(T) - \targetgait(\targetgaitpar)} + 
        \norm{\dgcurve(T) - \dtargetgait(\targetgaitpar)}
    \end{equation}
    \item Initial conditions: This constraint forces an initial point of a gait-switching trajectory should ``smoothly'' start from the current gait cycle,
    \begin{equation}
        \gcurve(0) = \currentgait(\currentgaitpar) \text{ and } \dgcurve(0) = \dcurrentgait(\currentgaitpar).
    \end{equation}
\end{enumerate}

\subsection{Torque-optimal trajectory}
\begin{enumerate}
    \item Decision variables: $\targetgaitpar$, $T$, $\effort_0$, and $\covyank[0]$.
    \item Boundary value error and Initial conditions: same as its definition of acceleration-optimal trajectory.
\end{enumerate}

\section*{Acknowledgment}
We are particularly grateful for the helpful insight of Alejandro Cabrera and the Office of Naval Research for support via grant N00014-23-1-2171.

\bibliographystyle{ieeetr}
\bibliography{ref}

\end{document}